\pdfoutput=1

\documentclass[11pt]{article}

\usepackage{EMNLP2023}

\usepackage{times}
\usepackage{latexsym}

\usepackage[T1]{fontenc}

\usepackage[utf8]{inputenc}

\usepackage{microtype}

\usepackage{inconsolata}

\usepackage[notransparent]{svg}

\usepackage{graphicx}
\graphicspath{{Figures/}{Fig/}}
\DeclareGraphicsExtensions{.png,.PNG,.jpg,.jpeg,.pdf}

\usepackage{color}
\usepackage[notextcomp]{stix}
\definecolor{burntsienna}{rgb}{0.91, 0.45, 0.32}
\definecolor{arsenic}{rgb}{0.23, 0.27, 0.29}

\definecolor{verdigris}{rgb}{0.26, 0.7, 0.68}
\definecolor{palegold}{rgb}{0.9, 0.75, 0.54}
\definecolor{tearose}{rgb}{0.97, 0.51, 0.47}	

\usepackage{float}
\title{Domain Knowledge Graph Construction Via A Simple Checker}

\author{Yueling (Jenny) Zeng and Li-C. Wang \\
  University of California, Santa Barbara \\
  Department of Electrical and Computer Engineering, Santa Barbara, CA 93106-9560, USA \\
  \texttt{yuelingzeng@ucsb.edu, licwang@ucsb.edu} }

\begin{document}
\maketitle
\begin{abstract}
With the availability of large language models, there is a growing interest for
semiconductor chip design companies to leverage the technologies. For those
companies, deployment of a new methodology must include two important considerations:
confidentiality and scalability. In this context, this work tackles the
problem of knowledge graph construction from hardware-design domain texts. 
We propose an oracle-checker scheme to leverage the power of GPT3.5 
and demonstrate that the essence of the problem is in distillation of 
domain expert's background knowledge. Using RISC-V unprivileged ISA 
specification as an example, we explain key ideas and discuss 
practicality of our proposed oracle-checker approach.  
\end{abstract}

\section{Introduction}

The release of large language models (LLMs), e.g., \cite{brown2020language, foundation-models, openai2023gpt4},
has motivated many semiconductor
chip design companies to 
license the technology for in-house use and explore various applications. 
One interesting application area is for better organizing the
tremendous amount of text data
accumulated over the years. 
For a semiconductor company, these data can include specifications, 
test plans, bug reports, meeting notes, and so on. They 
are collected for generations of designs from various stages of a design cycle. 
There is a constant need to
find ways to improve the 
within-the-company
knowledge accumulation and sharing over those data. 

\begin{figure}[thb]
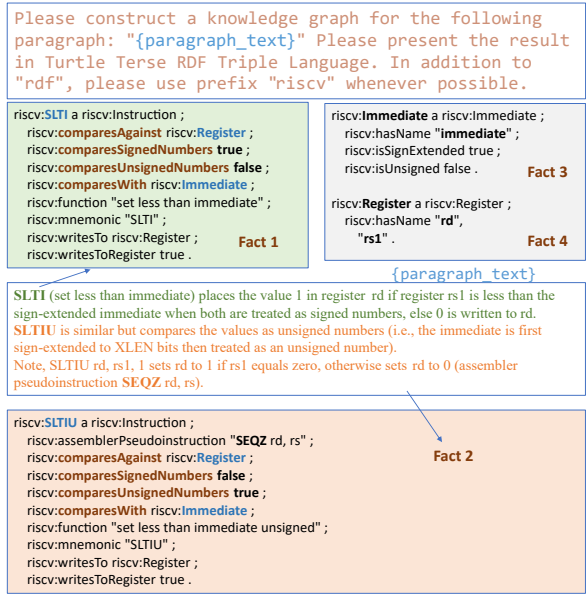

	\centering
	\includesvg[inkscapelatex=false, width=3in]{Figures/fig01a.svg}
	\includesvg[inkscapelatex=false, width=3in]{Figures/fig01.svg}
	\vspace{-0.1cm}
	\caption{\small An example of KG construction using a prompt to GPT3.5. 
		The input is a paragraph from the RISC-V ISA specification and the output
		is in RDF TTL format.}
	\label{fig01}
	\vspace{-0.1cm}
\end{figure}

In this context, we study the feasibility of using
an LLM for knowledge graph construction. 
There are two important considerations.
To protect proprietary information, the LLM has to be used in
an in-house environment closed from the outside world.
Because many of those companies might not maintain sufficient 
resources to conduct effective fine-tuning or retraining of an LLM,  
it is more practical to assume that the LLM would be used 
as it is. 
Furthermore, it is also desirable that the construction process of
knowledge graph can follow an iterative fashion, where
the graph is expanded gradually as more text items are processed. 

We therefore consider our knowledge graph construction (KGC) as the following: 
Given an ordered sequence of {\it text items} $T_1, \ldots, T_n$,
KGC processes one $T_i$ at a time from 1 to $n$ and generates an
individual knowledge
graph (KG): $g_i = KGC(T_i)$. Let $G_{i-1}$ be the KG after {\it merging} all
$g_1, \ldots, g_{i-1}$. We obtain $G_i = MERGE(G_{i-1}, g_i)$. 

In this work, we use RISC-V unprivileged ISA specification (``the Spec'') \cite{riscv-unpriviledged} as an example. 
We consider each paragraph as a text item. We choose GPT3.5 \cite{gpt3.5} as our LLM to use. 
For implementing the $KGC$ step, we rely only on prompting to the LLM. 

Figure~\ref{fig01} shows an example of the KGC. The KG is
represented in the Turtle Terse RDF Triple language (RDF TTL) \cite{ttl-format}. The specific
prompt in use is shown in the figure. 
The input paragraph can be seen in two parts: (1) 
definition of the SLTI instruction, and (2) 
definition of the SLTIU instruction. 

Notice that the second sentence starts with ``SLTIU is similar'' by 
referencing to the first sentence. The RDF output
is shown as four ``Facts''. The most interesting aspect of the result
is shown in ``Fact 2'' for SLTIU where the RDF duplicates all the
predicates used in ``Fact 1'', i.e. {\tt compareAgainst},
{\tt comparesSignedNumbers}, {\tt comparesUnsignedNumbers},
and {\tt comparesWith}. This indicates that the LLM does understand
the phrase ``is similar'' and reflects its understanding in copying
the RDF representation of SLTI.  

\subsection{Terminology}

We will use several terms in this paper to help discussion.
Refer back to Figure~\ref{fig01}. An RDF output is given as multiple {\it rdf blocks}.
We call each rdf block a {\it Fact}.  
Each Fact starts with a {\it subject entity}. For example, {\tt SLTI},
{\tt SLTIU}, {\tt Immediate}, and {\tt Register}, are subject entities.
A Fact represents a set of {\it triples} each in the form
$(subject, predicate, object)$. An {\it object entity} is the one that
appears as the object of a triple and is {\it not} a subject entity. 
We also differentiate two types of predicate. For a triple, if both of
its subject and object are subject entities, we call the predicate a
{\it relation}. Otherwise, we call it a {\it feature}. 
For example, in  Figure~\ref{fig01}  {\tt compareAgainst} is a relation
and {\tt comparesSignedNumbers} is a feature. 

For simplicity, we use the term ``RDF'' to refer to the RDF output given
by GPT3.5. 

\subsection{Background Facts (BFs)}

While Figure~\ref{fig01} shows some encouraging result, we notice that
the RDF misses some detail in the original paragraph. For example, 
it does not differentiate the usages of the two registers ``rd'' and ``rs1''.

\begin{figure}[thb]
	\centering
	\includesvg[inkscapelatex=false, width=3in]{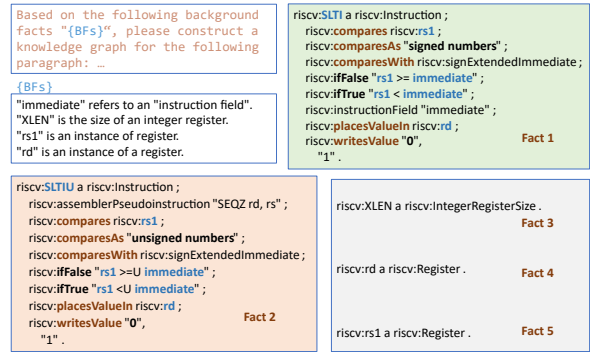}
	\vspace{-0.1cm}
	\caption{\small Improved RDF obtained by repeating the example in Figure~\ref{fig01}
		and by supplying {\it background facts} (BFs).}
	\label{fig02}
	\vspace{-0.1cm}
\end{figure}

Figure~\ref{fig02} shows another RDF by using a different prompt. 
This prompt adds a list of {\it background facts} (BFs) we manually
created. The RDF shows an improvement. In particular,
the RDF shows that the instructions {\tt compares} register ``rs1''
with the {\tt signExtendedImmediate}, and {\tt placesValuein} ``rd''.
It even includes the detail that the comparison is ``$<$'' (less than). 

\subsection{Focus of The Work}

The example above show that GPT3.5 alone {\it can} be used for KGC 
and produce reasonably good result.
Adding BFs can help improve the result further. 
It looks promising. 
However, it turned out that without adding BFs, the example was 
one of the few easy cases we encountered. Others were not as easy. 
Without BFs, an output RDF could be unsatisfactory due to two reasons
(see examples in Appendix~\ref{App.01}): (1) the RDF failed
the syntactic check, and (2) the RDF passed the syntactic check
but either got some fact(s) wrong or entirely missed
the main fact(s) described in the paragraph. 
In our experiment, we estimated at least 70\% of the cases were in 
these two categories.  

Figure~\ref{fig02} shows that adding BFs can influence the
behavior of GPT3.5 for KGC. Then, it is interesting to see whether 
we can rely solely on adding BFs to reach a satisfactory RDF 
for every paragraph or not. 
Ideally, we would like to turn the KGC process for every hard
case into an easy case (like Figure~\ref{fig02}). 
This work studies if the idea is feasible. 

\subsection{Contributions}

In this work we show that the simple approach of 
calling GPT3.5 with BFs can be sufficient. 
The contributions can be viewed in three aspects:

\noindent
(1) We propose an Oracle-Checker (OC) scheme to utilize GPT3.5. 
	In this scheme, we restrict ourselves to use a {\it simple} checker for practicality
	reason. A simple checker means that the checker does not involve sophisticated
	techniques (e.g. syntactic or semantic analyses, embedding training, etc.). 
	A simple checker promotes practicality because it is
	easier for the field engineers to adopt. \\
(2) With our OC scheme, we show that GPT3.5 is sufficient for the KGC task. 
	The key is to ``program'' the GPT3.5's KGC process with proper BFs. 
	Because these BFs are manually prepared, the essence of the KGC process 
	can be seen as distillation of those BFs, i.e. the required \textit{domain knowledge}. \\
(3) Through experiments, we observe and summarize findings regarding the
	GPT3.5's KGC process. Our findings can help others
	who desire to use GPT for a similar domain KGC task.

\section{Background and Related Work}

As far as we know, we are among the first to tackle the problem of KGC for 
unstructured text data in the semiconductor chip design domain. Documents in such a
company often involve terminologies not known to the outside world.
Hence, it is intuitive to think that KGC for those texts requires a
domain person to provide some background knowledge to at least cover those
terminologies. 
This thinking motivated us to take the view of supplying BFs. 

Knowledge graph construction is a rich field 
with many techniques having been proposed to {\it solve} 
the problem \cite{Yan2018ARO,ye-etal-2022-generative}.
Conventional methods to constructing knowledge graphs follow
a pipeline of NLP sub-tasks \cite{luan-etal-2018-multi} such as 
entity recognition \cite{tjong-kim-sang-de-meulder-2003-introduction}, 
entity linking \cite{entity-linking}, relation extraction \cite{zelenko-etal-2002-kernel}, 
and coreference resolution \cite{zelenko-etal-2004-coreference} etc. 
Among the various tasks, named entity recognition (NER) provides 
a fundamental first step for domain knowledge acquisition. 
Standard off-the-shelf toolkits for NER \cite{bird-2006-nltk, manning-etal-2014-stanford, finkel-etal-2005-incorporating, liao-veeramachaneni-2009-simple} 
combines machine learning models and rule-based components to label entities. 
Recent works solve the problem in an end-to-end fashion 
using deep learning models \cite{mondal-etal-2021-end, Harnoune_2021, ye-etal-2022-generative}.
However, pre-trained models often incur low accuracy  
since training data from general public domain rarely 
covers our domain specific patterns. 
It is hard to iteratively ingest domain experts' knowledge 
to further improve the accuracy without dedicated retraining.
Finetuning or retraining is often not a desirable option
for many hardware companies because of the tremendous efforts
required to create curated databases for the training tasks. 
Another potential route to our KGC problem is to implement 
customized rule-based extractors 
with features provided by existing constituency and/or dependency parsers 
\cite{chen-manning-2014-fast, zhu-etal-2013-fast}. 
However, to our experience the set of rules can 
quickly grow overly-complicated 
and it is difficult for the approach to scale. 

Restricting the output of a generative LM into a formal representation is related 
to the problem of constrained semantic parsing \cite{shin-etal-2021-constrained, lu-etal-2021-text2event}.
A structured meaning representation is often chosen as the output format \cite{wolfson-etal-2020-break}.
However, converting the meaning representation into a KG
can be another potential barrier. In summary, 
prompting an LLM to directly generate an RDF, if practical,
can bypass all the complications mentioned above. 
This would not be feasible without the latest developments of 
the LLMs. 

Despite that automatic knowledge graph construction in specific domain still 
remains an open challenge problem \cite{Yan2018ARO}, 
a major difference of this work is that 
we are not trying to propose another KGC solver. Rather, we focus on
verifying the result given by such a solver, in our case the GPT3.5. 
In other words, our work is {\it not} about being a technology provider
as those surveyed in \cite{Yan2018ARO, ye-etal-2022-generative}. Instead, we take
the perspective of being a technology consumer. From this perspective,
we study what features are required for a practical KGC solver.

\section{The Oracle-Checker View}

Our Oracle-Checker (OC) view was inspired by the theoretical model
of Interactive Proofs (IP) \cite{Arora:12}. An IP system
comprises a prover and a verifier. The prover is assumed to be
an all-powerful machine. The verifier is a probabilistic machine. 
The IP approach was developed to characterize 
computational complexity classes. 

In an IP system, the verifier interrogates the prover through a
sequence of communications. At the end the verifier either is 
convinced that the answer provided by the prover is correct or
reject it. An important aspect in the communications is to
keep the prover honest. Because verifier's computational power
is limited, it is mostly prover's job to make the verification task
as easy as possible. 

\begin{figure}[thb]
	\centering
	\includesvg[inkscapelatex=false, width=2.8in]{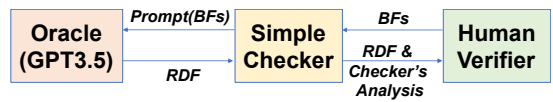}
	\vspace{-0.2cm}
	\caption{\small The proposed Oracle-Checker scheme}
	\label{figA}
	\vspace{-0.2cm}
\end{figure}

Our OC view is different from the IP approach though. This is  
because what we consider the
oracle (corresponding to the prover) has a limited power in practice.
In addition, our oracle is probabilistic. 
On the other hand, our checker still needs a way to
keep the oracle ``honest'', i.e. a way to verify
the answer provided by the oracle. However, because our oracle is
not all powerful, we can no longer expect the oracle to give a form
of answer that is always easily verifiable. This makes the verification
harder than that in the theoretical IP model. 

Figure~\ref{figA} depicts our OC scheme. In our scheme we introduce
a {\it human verifier}. The {\it verifier} comprises two parts: the simple checker
and the human verifier. The ultimate decision to accept or reject an
answer (RDF) stays with the human verifier. The job of the
simple checker is to analyze the answer and provide feedback 
to help the human verifier. 

Instead of asking the oracle to make the verification 
task easier for the verifier, if needed
we require the verifier to make the task
easier for the oracle. In our OC scheme, there are only two 
ways the verifier can do this: 
(1) by providing BFs and (2) occasionally by splitting
a paragraph into multiple sub-paragraphs. Note that making the task easier is opposite
to that in the theoretical IP model where the task is made easier
for the verifier. Therefore, we can think that in our OC scheme, 
the oracle is powerful but not all powerful,
and some of the power still resides with the human verifier. 

\subsection{Basic Requirements for Being An Oracle}

We impose two requirements for an LLM to be used as an oracle
in our OC scheme.
First, the LLM needs to have the ability to perform a {\it validity check}
on the answer it provides. 
For each RDF Fact, our simple checker asks the LLM
to perform an {\it entailment check}, asking whether or not the Fact can be
logically entailed from the paragraph (and if BFs are provided, with the BFs as well). 
If this check passes for every Fact in the RDF, the checker accepts. 
Otherwise, it rejects. Then, the checker reports the result to the
human verifier for review. 

The second requirement is that the oracle must be able to demonstrate
a {\it systematic} behavior in $N$ repeated runs. Because of the probabilistic
nature of an LLM, it is possible that in repeated runs, no two answers
are exactly the same. In this case, we consider the LLM failing 
the systematic requirement. In our experiment, 
if the LLM could produce at least two exact 
same answers in 10 repeated runs, we consider it satisfying the 
systematic requirement for the given task. 

\section{Feasibility Study}

We focus our discussion with paragraphs from the first two chapters
of the RISC-V Spec. The first chapter provides a general introduction. 
The second chapter provides specification of the instructions
from the RV32I integer instruction set. The rest of the
chapters are similar to the second chapter, providing specification
for a particular instruction set defined in RISC-V. The example in
Figure~\ref{fig01} is from chapter 2. Because the descriptions
from chapter 1 are more high-level, we expected that KGC would be more
difficult for those paragraphs. However, as our analysis will
show later, we find no significant difference on the 
GPT3.5's performance for paragraphs from the two chapters. 

\subsection{Consistency Check}
\label{sec04.2}

For checking the systematic requirement, included in our simple checker is a
{\it consistency check}. We repeat the same prompt 10 times and check
to see if at least two RDFs are exactly the same. Before checking
consistency, we also implement an RDF syntactic check using a
publicly-available RDF parser \cite{rdf-parser}. 
If an RDF fails the syntactic check, it is excluded for the
consistency check. 

\begin{figure*}[thb]
	\centering
	\includesvg[inkscapelatex=false, width=3.12in]{Figures/fig09.svg}
	\includesvg[inkscapelatex=false, width=3.12in]{Figures/fig10.svg}
	\vspace{-0.5cm}
	\caption{\small Results of consistency check without BFs provided; {\color{burntsienna} 
			$\mdblksquare$}: Most Consistent Group;  
		{\color{arsenic} $\mdblksquare$}: Failed}
	\label{fig09}
	\vspace{0.2cm}
	\centering
	\includesvg[inkscapelatex=false, width=3.12in]{Figures/fig11.svg}
	\includesvg[inkscapelatex=false, width=3.12in]{Figures/fig12.svg}
	\vspace{-0.5cm}
	\caption{\small Results of consistency check with BFs provided; {\color{burntsienna} 
			$\mdblksquare$}: Most Consistent Group;  
		{\color{arsenic} $\mdblksquare$}: Failed}
	\label{fig11}
	\vspace{-0.1cm}
\end{figure*}

For paragraphs in the two chapters, Figure~\ref{fig09} shows
the results of consistency check as two bar charts, for chapters 1 and 2 respectively. 
These results were obtained without BFs. 
The result of each paragraph 
may comprise multiple colored bars. Each color represents a group
of RDFs that are exactly the same. A dark bar
({\color{arsenic} $\mdblksquare$}) shows those
runs failing the syntactic check. Each orange bar
({\color{burntsienna} $\mdblksquare$}) denotes the 
largest group of RDFs that are exactly the same
(i.e. the most consistent group). 

Below some of the bars, there are text notes. Each note
means that for the original paragraph, multiple KGC trials failed
in all 10 runs. Then, we split the paragraph into multiple sub-paragraphs
to be processed separately. For example the first note is ``P10(3)''
indicating that paragraph 10 was split into 3 sub-paragraphs
in the experiment. 
We will discuss this splitting strategy later. However, notice that
some of the sub-paragraphs still fail the syntactic check even
after the splitting.  

Figure~\ref{fig09} demonstrates that in general GPT3.5 does have a
systematic behavior for KGC. For those failing cases, we then rely
on using BFs to resolve them. It is important to note that this
consistency check says nothing about the quality of the resulting RDFs. 
This assessment is done afterwards. 

\subsection{The Effect of Providing BFs}
\label{sec04.3}

Figure~\ref{fig11} shows the result of consistency check after
BFs are provided. For the two chapters, we had in total 204 BFs
(see Appendix~\ref{App.08}). 
In Figure~\ref{fig11}, there is no paragraph with a complete fail
any more. In the worst case, we obtained two RDFs that
are exactly the same. Based on the results, we can choose the
RDF from the most consistent group (the 
{\color{burntsienna} $\mdblksquare$} group)
to perform the entailment check. 

\subsection{Entailment Check}

\begin{figure}[h]
	\centering
	\includesvg[inkscapelatex=false, width=2.6in]{Figures/figB.svg}
	\vspace{-0.1cm}
	\caption{\small Two prompts used in the entailment check }
	\label{figB}
	\vspace{-0.1cm}
\end{figure}

Each entailment check is carried out with two prompts. 
The first prompt (Prompt A in Figure~\ref{figB}) asks GPT3.5 to
convert a Fact into a sentence. The second prompt (Prompt B) then asks GPT3.5
whether or not the given paragraph (and background facts if available) 
logically entails the statement of fact. In Figure~\ref{figB}, Prompt B is combined
with query 1 or query 2 to form two different prompts, one without BFs
and the other with BFs. 
It should be noted that given an RDF, the entailment check is applied
to each Fact individually. 
Recall that a Fact is an rdf block that may include multiple triples.

\begin{figure*}[h]
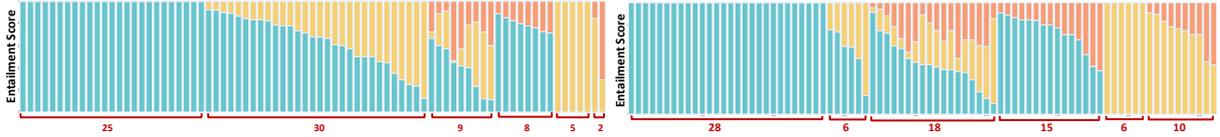

	\centering
	\includesvg[inkscapelatex=false, width=3.12in]{Figures/fig13.svg}
	\includesvg[inkscapelatex=false, width=3.12in]{Figures/fig14.svg}
	\vspace{-0.5cm}
	\caption{\small Results of entailment check for the two chapters of paragraphs;
		Each chart shows overlapping of two results from the runs without
		and with BFs provided; 
		{\color{verdigris} 
			$\mdblksquare$}: showing \% of RDF Facts passing the check
		where those Facts are obtained with no BFs provided;  
		{\color{palegold} 
			$\mdblksquare$}: showing \% of RDF Facts passing with BFs provided;  
		{\color{tearose} 
			$\mdblksquare$}: With BFs provided, some Facts fail the check 
		(mostly because of including the auxiliary entities not
		given in the paragraph) and
		are bypassed after manual review.  }
	\label{fig13}
	\vspace{-0.2cm}
\end{figure*}

Figure~\ref{fig13} summarizes the entailment check results for paragraphs
from the two chapters separately. 
The vertical axis shows the {\it entailment score}, a value
between 0 and 1:  
Assuming an RDF contains $N$ Facts. The ratio between the number of passing Facts
and $N$ is used as the score. 

Each bar in Figure~\ref{fig13} corresponds to the result from one paragraph
and can include three colors. The {\color{verdigris} $\mdblksquare$} color
bars are based on the RDFs obtained without BFs. 
If a bar has only this color, it means that adding BFs is not
necessary for passing the check. 
We may consider them as the ``easy'' cases. 

For those cases where the {\color{verdigris} $\mdblksquare$} color
bars do not reach the score 1.0, we then rely on BFs for bringing
the entailment check score to 1.0. 
For those that do not show up with a {\color{verdigris} $\mdblksquare$}  color
bar at all, they are the ``hard'' cases. 
Without BFs, there is no Fact passing the entailment check.

The {\color{palegold} $\mdblksquare$} bars then correspond to the
entailment scores based on the RDFs obtained with BFs provided. 
Note that these bars are shown behind the 
{\color{verdigris} $\mdblksquare$} bars. Consequently, if an original 
{\color{verdigris} $\mdblksquare$} color bar already reaches the
score 1.0 or if the {\color{palegold} $\mdblksquare$} bar is shorter
than the {\color{verdigris} $\mdblksquare$} bar, then the {\color{palegold} $\mdblksquare$} bar cannot be seen. 

For some paragraphs, we need the {\color{tearose} $\mdblksquare$} bars to
bring the score to 1.0. They represent Facts reported by the
simple checker as failing the entailment check after BFs are provided.
However, after manual review these failures are bypassed. 
Those failures can be divided into three categories where the first
one happens most frequently and the third happens only on very few cases.
See Appendix~\ref{App.02} for some examples
to illustrate these three categories. 

The first category is the creation and use of {\it auxiliary entities}
in the RDF. An auxiliary entity is the one that does not appear
in the paragraph (nor the BFs) and is created to facilitate
describing other entities. In a sense, we can consider those 
auxiliary entities as BFs automatically supplied by GPT3.5. 
Because they are not mentioned in the paragraph and the BFs,
Facts involving them would fail the entailment check. 
However, the GPT3.5's ability to add auxiliary entities 
(i.e. its own BFs) can be quite desirable, because it 
helps provide BFs possibly missed by our manual preparation 
of BFs. 

The second category involves an entity or a predicate that
has nothing wrong by itself. However, in the RDF the entity/predicate
is specified within a particular namespace (e.g. ``riscv:'', ``rdf''). 
Because the original paragraph (and the BFs) do not explicitly
state their use in the namespace, this can cause the entailment
check to fail. 

The third category, happening only on few cases, involves 
the use of namespaces other than rdf or risvc namespaces.
In the Spec, there are some descriptions about other ISA 
architectures (MIPS, SPARC, etc.). Those descriptions may result
in the creation of their respective namespaces. In an RDF, entities
from two namespaces might be connected through a predicate. 
This can cause a problem for the entailment check as the original
paragraph (and the BFs) simply considers those terms as entities
rather than different namespaces. 


In summary, 
Figures~\ref{fig09}-\ref{fig13} shows that it is feasible to use
GPT3.5 as an oracle. Further, providing BFs can improve
consistency and also help reach a satisfactory RDF 
for every paragraph. 

\begin{figure}[h]
	\centering
	\includesvg[inkscapelatex=false, width=1.8in]{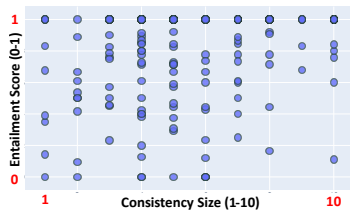}
	\vspace{-0.1cm}
	\caption{\small No correlation between the size of largest group
		from consistency check and the entailment check score}
	\label{fig16}
	\vspace{-0.3cm}
\end{figure}


It should be noted that more consistency does not mean more likely to
pass the entailment check. Figure~\ref{fig16} illustrates this point.
The x-axis is the size of largest consistent group. The y-axis is 
the entailment score. Every dot is a paragraph. The RDF is the one
obtained without BFs provided (i.e. Figure~\ref{fig11}). As seen,
the results
of consistency check and of entailment check have no obvious correlation.

\section{Analysis of Results}

We let BF$\phi$ refer to the KGC without
BFs, and BF$A$ refer to the KGC with BFs added. 
Further, BF$\phi$-Pass refers to those paragraphs whose RDFs pass the entailment
check, and BF$\phi$-Fail refers to those failing the check. 
Next we focus on comparing the the changes
of RDFs from BF$\phi$ to BF$A$.

There can be several metrics to measure the RDF changes from
BF$\phi$ to BF$A$, including: (1) the number of entities 
({\it entity coverage}), (2) the set of subject entities,
(3) the number of triples ({\it fact coverage}), and (4)
the number of predicates (relations or features or both).
In addition, we can consider those metrics separately from 
Facts passing and from Facts failing the entailment check. 

From the KGC process, in addition to consistency we can
consider another metric measuring the variations of 
subject entities across the 10 RDFs. We use a {\it conformity score}
defined as the ratio between the sum of 
individual numbers of subject entities and the size of the union set 
of the subject entities, across the 10 RDFs. 
A score 10 means every RDF uses the same set of subject entities.
A score 1 means everyone uses a different set. 
We found no obvious trend based on the changes 
of conformity scores from BF$\phi$ to BF$A$. We also found no
obvious correlation between the conformity score changes
and the changes on the number of entities from BF$\phi$ to BF$A$
(see Section~\ref{App.03}).  

In general we found the number of predicates a less effective
metric. This is because the same predicate can be used in many
triples. Hence the number of predicates used in an RDF reveals
little information about each predicate's usage frequency. 

Note that for a given metric, separating its 
numbers based on passing and failing the entailment check can 
be difficult. This is because entailment check is applied to a Fact. 
As a result, pass/fail is applied to the entire Fact and may
not reflect the pass/fail of a particular triple, entity, or predicate.

We did find that separating the scenario
BF$\phi$-Pass from the scenario BF$\phi$-Fail was helpful,
enabling us to observe some interesting trends. 

\subsection{Entity Coverage}

\begin{figure}[h]
	\centering
	\vspace{-0.1cm}
	\includesvg[inkscapelatex=false, width=2.8in]{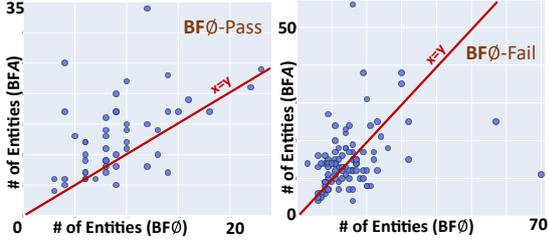}
	\vspace{-0.2cm}
	\caption{\small Comparing entity coverage from BF$\phi$ to BF$A$ for
	the two scenarios: BF$\phi$-Pass and BF$\phi$-Fail }
	\label{fig19-20}
	\vspace{-0.2cm}
\end{figure}

Figure~\ref{fig19-20} shows an interesting trend. 
For most of the cases in the BF$\phi$-Pass category,
the BF$A$ produces an RDF that contains
more number of entities than the corresponding BF$\phi$'s RDF, i.e. 
improving the entity coverage.
Intuitively, BF$A$ keeps most of the entities from BF$\phi$-Pass because 
Facts involving them pass the entailment check already
(they are ``correct'' and no need to replace them). 

In contrast, for those paragraphs in the BF$\phi$-Fail category,
the effect can go either ways, i.e. the entity coverage can increase
or decrease. 
The RDFs in the BF$\phi$-Fail category include 
Facts that fail. BF$A$ intends to fix those problems. 
This fixing can involve finding a different set of subject entities, 
resulting in using more or using less entities. 

\subsection{Fact Coverage}

\begin{figure}[h]
	\centering
	\includesvg[inkscapelatex=false, width=2.3in]{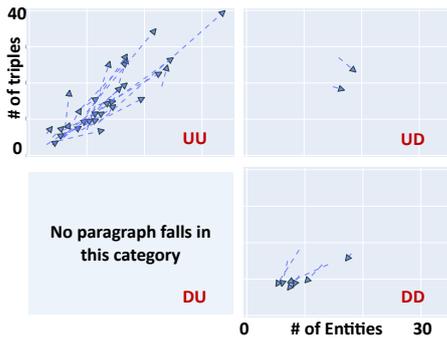}
	\vspace{-0.1cm}
	\caption{\small For those paragraphs where BF$\phi$ passes the entailment
		check, adding BFs generally results in not only more entities 
	but also more triples included in the RDF (the UU chart); The first U (D) stands for
    \# of entities going up (down) and the second U (D) stands for
    \# of triples going up (down)}
	\label{fig21}
	\vspace{-0.1cm}
\end{figure}

\begin{figure}[h]
	\centering
	\includesvg[inkscapelatex=false, width=2.3in]{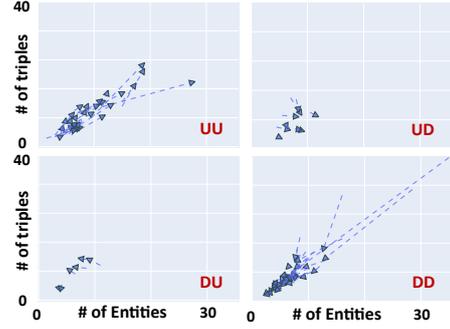}
	\vspace{-0.1cm}
	\caption{\small For those paragraphs where BF$\phi$ fails the entailment
		check, adding BFs generally results in RDFs falling into either the UU or
		the DD
		categories; There are several exceptions in the UD and DU categories	 }
	\label{fig22}
	\vspace{-0.1cm}
\end{figure}

Although we generally expect that using more entities would
result in more triples in the RDF, 
Figure~\ref{fig21} and Figure~\ref{fig22} show that
it is not always the case. 
These figures plot the number of entities against the
number of triples. An arrow shows the change from BF$\phi$
to BF$A$. 
Although we expect most cases fall into the UU
(both numbers increase) and DD (both numbers decrease)
categories, we do see cases in the other two categories. 

There are two cases shown on the UD chart in Figure~\ref{fig21}. 
The top arrow is associated with the paragraph discussed earlier
in Figure~\ref{fig01} and Figure~\ref{fig02}. 
The case associated with the second arrow is similar
(see Appendix~\ref{App.04} for more of its discussion). 

Like the BF$\phi$-Fail chart shown in Figure~\ref{fig19-20} where  
the changes from BF$\phi$ to BF$A$ can go both ways, Figure~\ref{fig22}
shows that the changes when considering the number of triples can
go four ways. Most still follows our expected trends as UU and DD.
However, there are quite a few in the UD and DU categories. 
As BF$\phi$'s RDFs failed the entailment check, the ways BF$A$ fixed
them can be diverse, resulting in the four different scenarios as seen.  

\subsection{Root Subject Entity (RSE) Carry-Over}

A {\it root subject entity} (RSE) is a subject entity that does not appear
in the object field of any triple in a given RDF. 
Root entities can be thought of as the main topics extracted
by the oracle. It is interesting to analyze how 
many root entities from BF$\phi$ are kept in the RDF of BF$A$. 
We use a percentage number (0 to 1) to capture the extent of this
RSE carry-over from BF$\phi$. 

\begin{figure}[hbt]
	\centering
	\includesvg[inkscapelatex=false, width=3.12in]{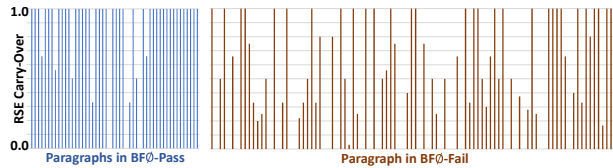}
	\vspace{-0.4cm}
	\caption{\small \% of root subject entities (RSEs) carried over from the 
		RDF of BF$\phi$
		into the RDF of BF$A$  }
	\label{fig23}
	\vspace{-0.1cm}
\end{figure}

Figure~\ref{fig23} shows the RSE carry-over, separated again by
the two scenarios, BF$\phi$-Pass and BF$\phi$-Fail. 
It can be seen that for majority of cases in the BF$\phi$-Pass
category RSE carry-over is 100\%.
This means that all RSEs used in the BF$\phi$'s RDF are kept
in the RDF of BF$A$ (BF$A$ may contain many other RSEs though).
For those cases, what the BF$A$ primarily doing is to expand the fact coverage, i.e.
finding additional RSE and generating more triples. 

In addition to 100\% carry-over, the other two cases (partly and no
carry-over) can be due to various reasons. They generally indicate that
BF$A$ finds a different set of main topics for the KGC. 

On the BF$\phi$-Pass chart in Figure~\ref{fig23}, there are four
cases where the RSE carry-over is 0. Their reasons vary. 
Appendix~\ref{App.05} 
provides their details for further reference. 

As for the cases in BF$\phi$-Fail, the effects of adding BFs
can be diverse. Appendix~\ref{App.06} provides
two examples to illustrate the effects.  

\subsection{Merging Individual RDFs}

With BFs added, RDFs obtained by BF$A$ for individual paragraphs would share
more entities that enable more connections among the paragraphs.
This finding is illustrated with additional analyses in Appendix~\ref{App.07}.  
From the perspective of merging individual RDFs into a single KG,
it is interesting to note that before BFs was provided,
the RDFs from BF$\phi$ already had many shared entities
(if we do not consider whether or not they pass the entailment check). 
BFs further added more sharing to the resulting RDFs. 
With this observation, we see that the primary objective to add BFs
is for fixing the problems seen in the RDFs from BF$\phi$.
Improving entity sharing becomes secondary.

\section{The Tasks of Human Verifier}

There are two primary tasks for the human verifier: (1) deciding
when to split a paragraph into sub-paragraphs, and (2) deciding what BFs
to provide. 

Empirically, splitting a paragraph can be due to three reasons: 
(1) When a paragraph is too lengthy (e.g. a long paragraph followed by a list of
bullet points), splitting can help. This is the obvious case. 
(2) If a paragraph contains
a question raised, followed by an answer author provides, then it is
helpful to separate the question from the answer. 
(3) When a paragraph contains statements regarding other architectural
domains (e.g. MIPS, SPARC, etc), it is helpful to separate those
non-RSIC-V descriptions from the RISC-V descriptions. 

To our experience, the results from BF$\phi$ can be informative
for choosing BFs. Further, preparing BFs can follow
several guidelines: (1) For an abbreviation or convention defined
earlier or somewhere else, a BF needs
to be provided. If not, GPT3.5 may
supply its own meaning which can be wrong. (2) When we want 
to guide the KGC to focus on an entity as the main topic, 
we can provide a BF about the entity (see the example
in Section~\ref{App.06.1}). (3) When the paragraph involves
several concept described in long phrases, providing 
BFs that emphasize each long phrase as a single concept
can help reduce the KGC complexity (see the example in Section~\ref{App.06}). 
(4) When we desire
to include a missing fact detail involving an entity, we can ``remind''
GPT3.5 by providing a BF about the entity (e.g. Figure~\ref{fig02}).
However, this might not always work as occasionally GPT3.5 can 
choose ignore a BF completely. 

In general, we found that adding too many unnecessary BFs could
degrade the quality of the resulting RDF. Hence, the BFs
should not be excessive. We also found that we should avoid
adding a BF describing a relation between two entities
which are already related by the paragraph. 
This superimposition of relations could cause the KGC process
to fail or even enter a repeated loop. 

\section{Limitations}

As the first step, our work shows the feasibility of
the proposed OC scheme. We have not demonstrated how to
improve its automation. Manual efforts are 
required to prepare the BFs and review the entailment check results. 
Although we have seen several
guidelines to prepare BFs, the manual process remains ad hoc
at this stage. 

Note that with more familiarity of the content described in 
the Spec, we found that preparing the BFs could be done quickly
(minutes for a section). Reviewing the results from the entailment
check was not a time-consuming task either. The most time-consuming
task was to deal with those difficult cases when the RDF remained
unsatisfied after several trials. We could spend over an hour
on each of those cases (counting the GPT3.5 calling time).  

\section{Conclusion and Future Work}

This work demonstrates the feasibility to use GPT3.5 
for KGC on texts in the semiconductor chip design domain. 
We propose an OC scheme to use GPT3.5 as an
oracle and the essence is using BFs
to influence the oracle's behavior. The
communications to GPT3.5 are through a simple checker
that imposes two checks: the consistency check and the
entailment check. We consider them 
as requirements for an LLM to be used as an oracle. 

Our current OC scheme requires a human verifier whose
primary job is to review the results of the entailment check
and prepare BFs. To what extent this
manual process can be automated and how to do it 
are interesting questions for further research.

\bibliography{anthology,custom}
\bibliographystyle{acl_natbib}

\newpage

\appendix
\section{Appendix}
\label{sec:appendix}

\subsection{Examples of Difficult Cases Without BFs}
\label{App.01}

We use two examples to show that not all paragraphs are as easy to handle
by GPT3.5 as the paragraph
used in Figure~\ref{fig01} and Figure~\ref{fig02}. 

\subsubsection{The RDF Failing The Syntactic Check}
\label{App.01.1}

The paragraph of the first example is shown in Figure~\ref{figAppExample1a}.
Similar to the paragraph shown in Figure~\ref{fig01}, this paragraph was
also about defining several instructions. It turned out that this paragraph caused
some issue in the RDF syntactic check. Without BFs provided, all RDFs from 10 repeated
trials failed the syntactic check
(The problem came from its choice to use ``XORI rd, rs1, -1'' as a single entity).  

\begin{figure}[thb]
	\centering
	\includesvg[inkscapelatex=false, width=3in]{Figures/figAppExample1a.svg}
	\vspace{-0.3cm}
	\caption{\small A problematic example for GPT3.5 to construct
		the KG using the same prompt in Figure~\ref{fig01} without 
		using any background fact, even though the paragraph
	is also about defining instructions like the one in Figure~\ref{fig01} }
	\label{figAppExample1a}
	\vspace{-0.1cm}
\end{figure}

We then provided the BFs as shown in Figure~\ref{fig02}
for ``immediate'', ``rs1'' and ``rd''. 
We used the same prompt as that shown in Figure~\ref{fig02}. 
The resulting PDF is shown in Figure~\ref{figAppExample1b}. 
The paragraph defined
four instructions covered by four Facts in the RDF. 

\begin{figure}[thb]
	\centering
	\includesvg[inkscapelatex=false, width=2.8in]{Figures/figAppExample1b.svg}
	\vspace{-0.1cm}
	\caption{\small The RDF obtained for the paragraph in Figure~\ref{figAppExample1a}}
	\label{figAppExample1b}
	\vspace{-0.1cm}
\end{figure}

\subsubsection{Unsatisfactory RDF}
\label{App.01.2}

Figure~\ref{fig03} shows another paragraph that is difficult
for GPT3.5 to handle. In repeating 10 requests using the same
prompt as that in Figure~\ref{fig01}, only one succeeded with 
a RDF that passes the syntactic check. 
The one result is shown in Figure~\ref{fig04}. 

\begin{figure}[thb]
	\centering
	\includesvg[inkscapelatex=false, width=3in]{Figures/fig03.svg}
	\vspace{-0.1cm}
	\caption{\small A difficult example for GPT3.5 to construct
		the KG using the same prompt in Figure~\ref{fig01} without 
		supplying any background fact. The main subject entities
		of the paragraph are the four ISA variants highlighted
		(RV32I, RV64I, RV128I, RV32E). A good
		KG constructor is required to uncover their relations and the
		features to describe their differences.}
	\label{fig03}
	\vspace{-0.1cm}
\end{figure}

The RDF was not satisfactory because the paragraph says that the four
variants were differentiated by two aspects: the size of the address space
and the number of integer registers. The RDF represented the second aspect
explicitly but did not mention anything about the address space. 
Furthermore, the paragraph did not say that RV64I and RV128I both had
32 integer registers like RV32I. 

\begin{figure}[thb]
	\centering
	\includesvg[inkscapelatex=false, width=2.5in]{Figures/fig04.svg}
	\vspace{-0.1cm}
	\caption{\small One RDF output given for the paragraph in Figure~\ref{fig03}
		using the same prompt as that in Figure~\ref{fig01}}
	\label{fig04}
	\vspace{-0.2cm}
\end{figure}

Using the same prompt in Figure~\ref{fig02}, we used the 
BFs in Figure~\ref{fig06} for the paragraph shown in Figure~\ref{fig03}.
The most consistent RDF is shown in Figure~\ref{fig05}.
The consistency check was explained in Section~\ref{sec04.2}).

\begin{figure}[thb]
	\centering
	\includesvg[inkscapelatex=false, width=2in]{Figures/fig06.svg}
	\vspace{-0.1cm}
	\caption{\small BFs provided to obtain result in Figure~\ref{fig05}}
	\label{fig06}
	\vspace{-0.1cm}
\end{figure}

Comparing to the result in Figure~\ref{fig04}, the new RDF captured
many more facts (i.e. triples). Notice that the four subject entities
Were the same between the two RDFs. The new RDF included
the triple ``(RV32E, subClassOf, RV32I)'' which was missing previously. 
This was a correct relation between the 32E variant and the base 32I. 

The new RDF also included the entity {\tt addressSpace} which
was missing in the previous result. Instead of using the long
predicate {\tt has\_Number\_of\_Integer\_Resgiters} as that in Figure~\ref{fig04},
the new RDF used a less specific entity {\tt integerRegister} and
represented its length with {\tt XLEN} in bits (as it should). It also
included a comment stating that
32E had half the number of integer registers. Overall, the new RDF
included many more correct facts from the paragraph. 

\begin{figure}[thb]
	\centering
	\includesvg[inkscapelatex=false, width=3in]{Figures/fig05.svg}
	\vspace{-0.2cm}
	\caption{\small The RDF result for the Figure~\ref{fig03} example }
	\label{fig05}
	\vspace{-0.1cm}
\end{figure}

\begin{figure}[thb]
	\centering
	\includesvg[inkscapelatex=false, width=2in]{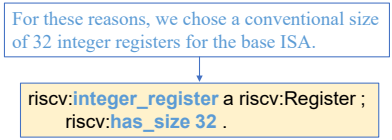}
	\vspace{-0.1cm}
	\caption{\small The number of integer registers is explicitly 
		stated in a sentence later separated by 62 paragraphs from the paragraph
		shown in Figure~\ref{fig03}}
	\label{fig07}
	\vspace{-0.1cm}
\end{figure}

It was interesting to note that the paragraph in Figure~\ref{fig03}
did not explicitly state the number of registers for the four base ISAs. 
This was why the RDF in Figure~\ref{fig05} did not include this information. 
In the Spec, this information was actually stated in another paragraph 
62 paragraphs later. Figure~\ref{fig07} shows the sentence
and the corresponding rdf block. 

The two RDFs in Figure~\ref{fig05} and Figure~\ref{fig07} can be connected
through the same entity referring to the ``integer register.'' Because in our
KGC each paragraph (i.e. ``a text item'') is handled separately, including a 
more generic entity in the RDF is crucial from the perspective of merging
individual RDFs into a whole knowledge graph. From this perspective, 
Figure~\ref{fig04} is less desirable even if it were correct.

\subsection{Examples Failing The Entailment Check}
\label{App.02}

For the two charts shown in Figure~\ref{fig13}, we used six
examples (three from each chart) to illustrate what happened to
those Facts represented by the {\color{tearose} $\mdblksquare$} bars. 
On each chart, those {\color{tearose} $\mdblksquare$} bars
appeared in three of the last four groups from the right. 
For each group, we selected the first paragraph from the right
as the example for the group. 
We started with the first paragraph from the right in the second chart
as the first example, and moved left to the other five examples. 
On each example, we showed the paragraph followed by
the Facts that failed the entailment check. 

\subsubsection{First Example}

\begin{figure}[h]
	\centering
	\includesvg[inkscapelatex=false, width=3in]{Figures/figAppP79.svg}
	\vspace{-0.1cm}
	\caption{\small The three Facts fail the entailment check
	because of the {\tt ISA\_property} created}
	\label{figAppP79}
	\vspace{-0.1cm}
\end{figure}

The paragraph of the first example mentioned three types of impacts:
on the ``code size'', on the ``performance'', and on the ``energy
consumption''. In the RDF, those three impacts were treated as three
separate subject entities. However, in order to group them together
a separate entity {\tt ISA\_property} was created. The entailment
fails because the paragraph did not mention specifically that
the RISC-V domain involved an entity called {\tt ISA\_property}. 
In other words, {\tt ISA\_property} can be seen as an auxiliary
entity (or concept) created to facilitate organization of other 
facts in the RDF. 
Auxiliary entities are not explicitly specified and hence, their use 
can fail the entailment check. 

From a practical point of view, we consider automatic creation of 
auxiliary entities and their proper use
a desirable capability of GPT3.5 for KGC. Because of this capability,
we do not need to explicitly state all BFs that might be used to
interpret and/or organize the entities (concepts) 
discussed in a given
paragraph. For example, we might provide the BFs explicitly stating
that the three different impacts are a type of some property to 
facilitate their organization. However, we did not need to 
because GPT3.5 automatically created them. 

\subsubsection{Second Example}

\begin{figure}[h]
	\centering
	\includesvg[inkscapelatex=false, width=3in]{Figures/figAppP112.svg}
	\vspace{-0.1cm}
	\caption{\small In this example, Facts 1--3 failed
		the entailment check because the auxiliary
		entities {\tt JALInstruction}, {\tt ImmediateType} and {\tt PipelineConcept}
		created, respectively. Additionally Fact 2 failed because the use of
		the rdf predicate {\tt subClassof}.}
	\label{figAppP112}
	\vspace{-0.1cm}
\end{figure}

The reason for the second example failure was similar to the first
example discussed above. In addition, the entailment check complained
about the use of the rdf predicate {\tt subClassof}, i.e. there was no
specific definition about the particular class hierarchy in the RISC-V
domain. Again, using the rdf predicate {\tt subClassof}
to organize entities (concepts) into a hierarchy can actually be 
helpful. During our manual review, if an entailment check failed
because of this reason, we usually ignored it. 

\subsubsection{Third Example}

\begin{figure}[h]
	\centering
	\includesvg[inkscapelatex=false, width=3in]{Figures/figAppP101.svg}
	\vspace{-0.1cm}
	\caption{\small In this example, Fact 1 failed
		the entailment check because the auxiliary
		entity {\tt BranchTarget}.}
	\label{figAppP101}
	\vspace{-0.1cm}
\end{figure}

For the third example, the reason of failing the entailment check
was the same as that in the previous two examples. 
It was interesting to observe that the auxiliary entity 
{\tt BranchTarget} was created to classify the term ``overflow''.
Indeed, the ``overflow'' was the branch target of the 
``blt'' instruction in this case. 

\subsubsection{Fourth Example}

\begin{figure}[h]
	\centering
	\includesvg[inkscapelatex=false, width=3in]{Figures/figAppP23.1.svg}
	\vspace{-0.1cm}
	\caption{\small In this example, the Facts fail because of
	the various namespaces (``mips:'', ``sparc:'', and ``isa:'') in use}
	\label{figAppP23.1}
	\vspace{-0.1cm}
\end{figure}

The fourth example shows that entailment check can fail due to
an auxiliary namespace created. The RDF in Figure~\ref{figAppP23.1} 
shows two namespaces created: {\tt mips:} and {\tt sparc:}.
In order to organize these two namespaces, an auxiliary namespace
{\tt isa:} was used. The paragraph did not specify explicitly
the namespaces and the relations of terms in those namespaces. 
They were created to facilitate representation of the information. 
We observed in our experiment that when a fact involved a relation
across two namespaces, it can cause a failure in the entailment
check because the original paragraph did not concern about the
namespaces created (i.e. those names were simply ``concepts'',
not namespaces).

\subsubsection{Fifth Example}

\begin{figure}[h]
	\centering
	\includesvg[inkscapelatex=false, width=3in]{Figures/figAppP51.svg}
	\vspace{-0.1cm}
	\caption{\small This example shows the creation of another type
	of auxiliary entities which we can call them ``intermediate''
	subject entities. 
    These entities are not mentioned in the paragraph and
    are created to represent the intermediate concepts that can be used
    to describe other entities}
	\label{figAppP51}
	\vspace{-0.1cm}
\end{figure}

In the fifth example, four auxiliary subject entities
{\tt Address\_1}, {\tt Address\_2}, {\tt Bits\_1}, and
{\tt Bits\_2} were created to describe other entities. 
It was interesting to notice that these four entities corresponded
to the ``first halfword address'' (byte 0), ``second halfword address''
(byte 2), ``first halfword'' (bits 15-0), and ``second halfword
(bits 31-16), respectively. We can call them ``intermediate''
entities as they are created to represent intermediate concepts
used to describe other entities. These auxiliary entities
did not appear in the paragraph (nor in the BFs). Hence, entailment
check involving them would fail.

\subsubsection{Sixth Example}

\begin{figure}[h]
	\centering
	\includesvg[inkscapelatex=false, width=3in]{Figures/figAppP59.svg}
	\vspace{-0.1cm}
	\caption{\small This example shows that entailment check
	can fail due to not explicitly specifying a predicate in 
    the given domain. In this case, the {\tt refersTo} predicate is not
    explicitly specified as a predicate in the RISC-V domain}
	\label{figAppP59}
	\vspace{-0.1cm}
\end{figure}

In the sixth example, the BFs provided further information about the
term ``trap''. However, it did not explicitly define that 
``refers to'' was a predicate specific in the RISV-C domain. 
This type of domain restriction can cause the entailment check
to fail, i.e. the original statement talked about a general term
while the RDF restricted the term to be specific to the
RISC-V domain. Our manual review bypassed this type of entailment
check failures.

\subsection{Variations in Subject Entities}
\label{App.03}

\begin{figure}[h]
	\centering
	\includesvg[inkscapelatex=false, width=1.8in]{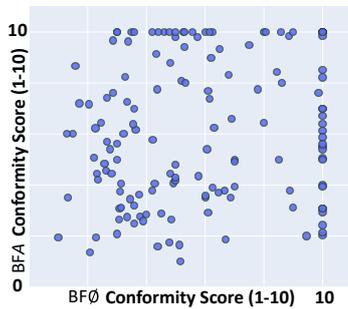}
	\vspace{-0.1cm}
	\caption{\small Conformity measuring the variations of subject entities across
		10 repeated runs: BF$\phi$ Vs. BF$A$}
	\label{fig17}
	\vspace{-0.1cm}
\end{figure}

Figure~\ref{fig17} shows that there is no obvious trend on the
changes of conformity scores from BF$\phi$ to BF$A$. Every dot in
this figure represents a paragraph.

\begin{figure}[h]
	\centering
	\includesvg[inkscapelatex=false, width=2in]{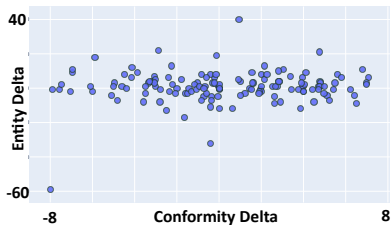}
	\vspace{-0.1cm}
	\caption{\small No correlation between the change of conformity score
		and the change of the number of entities after BFs added}
	\label{fig18}
	\vspace{-0.1cm}
\end{figure}

Figure~\ref{fig18} further shows that there is no correlation between
the changes of the conformity score and the changes of the number of
entities used in the RDF, as we move from BF$\phi$ to BF$A$. 

\subsection{Case Discussion from Figure~\ref{fig21}}
\label{App.04}

\begin{figure}[h]
	\centering
	\includesvg[inkscapelatex=false, width=3in]{Figures/figAppP82a.svg}
	\vspace{-0.4cm}
	\caption{\small The paragraph associated with one of the two
		arrows shown on the UD chart in Figure~\ref{fig21}}
	\label{figAppP82a}
	\vspace{0.3cm}
\end{figure}

For the two arrows shown on the DD chart in Figure~\ref{fig21}, the 
paragraph associated with one of the arrows is 
the one already explained
with Figure~\ref{fig01} and Figure~\ref{fig02}. The other paragraph is shown
in Figure~\ref{figAppP82a}. 
Figure~\ref{figAppP82b} shows the corresponding RDF from BF$\phi$ and
Figure~\ref{figAppP82c} shows the RDF from BF$A$. 

\begin{figure}[H]
	\centering
	\includesvg[inkscapelatex=false, width=2.8in]{Figures/figAppP82b.svg}
	\vspace{-0.1cm}
	\caption{\small The BF$\phi$'s RDF for the paragraph shown
		in Figure~\ref{figAppP82a}, containing 19 triples}
	\label{figAppP82b}
	\vspace{0.7cm}
	\centering
	\includesvg[inkscapelatex=false, width=2.8in]{Figures/figAppP82c.svg}
	\vspace{-0.1cm}
	\caption{\small The BF$A$'s RDF for the paragraph shown
		in Figure~\ref{figAppP82a}, containing 18 triples}
	\label{figAppP82c}
	\vspace{-0.1cm}
\end{figure}

As seen, the BF$A$'s RDF used more entities (including those 
represented as strings) but one less triple. 
The two RDFs essentially captured the same facts but
represented them differently. 

Notice that the provided BFs enabled 
BF$A$'s RDF to connect the particular exception concept, 
{\tt InstructionAddressMisalignedException},
to the hardware thread, {\tt Hart}. 
As the concept ``hart" was discussed in other paragraphs, this connection
facilitated merging of KGs between this paragraph and those other paragraphs.

\subsection{Four Cases With RSE Carry-Over $=0$}
\label{App.05}

There are four cases whose RSE carry-over is 0 on 
the BF$\phi$-Pass chart in Figure~\ref{fig23}. 
We provide their details in this section.
Figure~\ref{figApp4RSE-CO} shows the four paragraphs 
and the changes of entities in use from BF$\phi$
to BF$A$ (from left to right as pointed by the arrow). 
The BF$\phi$'s RDFs and BF$A$'s RDFs for the four cases
are then shown in Figures~\ref{figAppP20}, 
\ref{figAppP39}, (\ref{figAppP63a}, \ref{figAppP63b}), and \ref{figAppP81},
respectively.

\begin{figure}[H]
	\centering
	\includesvg[inkscapelatex=false, width=2.7in]{Figures/figApp4RSE-CO.svg}
	\vspace{-0.1cm}
	\caption{\small More details on the four cases whose RSE carry-over is 0 on 
		the BF$\phi$-Pass chart in Figure~\ref{fig23}}
	\label{figApp4RSE-CO}
	\vspace{-0.1cm}
\end{figure}

The first case was a single sentence. Even without the BFs provided,
GPT3.5 understood what the four base ISAs were for RISC-V. 
To represent the information that ``the four base ISAs were
distinct base ISAs'', the BF$\phi$'s RDF explicitly represented
the facts that they were mutually distinct. It used the relation
{\tt isDisinctFrom} for this purpose. 
The RSE was {\tt RISC-V} used to group the four base ISAs together.

\begin{figure}[htb]
	\centering
	\includesvg[inkscapelatex=false, width=3in]{Figures/figAppP20.svg}
	\vspace{-0.1cm}
	\caption{\small Comparison of BF$\phi$'s RDF and BF$A$'s RDF
	for the first case in the table shown in Figure~\ref{figApp4RSE-CO}}
	\label{figAppP20}
	\vspace{-0.1cm}
\end{figure}

\begin{figure}[htb]
	\centering
	\includesvg[inkscapelatex=false, width=3in]{Figures/figAppP39.svg}
	\vspace{-0.1cm}
	\caption{\small Comparison of BF$\phi$'s RDF and BF$A$'s RDF
		for the second case in the table shown in Figure~\ref{figApp4RSE-CO}}
	\label{figAppP39}
	\vspace{-0.1cm}
\end{figure}

\begin{figure}[htb]
	\centering
	\includesvg[inkscapelatex=false, width=2.8in]{Figures/figAppP63a.svg}
	\vspace{-0.1cm}
	\caption{\small BF$\phi$'s RDF for the third case in the table shown in Figure~\ref{figApp4RSE-CO}}
	\label{figAppP63a}
	\vspace{-0.1cm}
\end{figure}

\begin{figure}[htb]
	\centering
	\includesvg[inkscapelatex=false, width=2.8in]{Figures/figAppP63b.svg}
	\vspace{-0.1cm}
	\caption{\small BF$A$'s RDF
		for the third case in the table shown in Figure~\ref{figApp4RSE-CO}}
	\label{figAppP63b}
	\vspace{-0.1cm}
\end{figure}

\begin{figure}[htb]
	\centering
	\includesvg[inkscapelatex=false, width=3in]{Figures/figAppP81.svg}
	\vspace{-0.1cm}
	\caption{\small Comparison of BF$\phi$'s RDF and BF$A$'s RDF
		for the fourth case in the table shown in Figure~\ref{figApp4RSE-CO}}
	\label{figAppP81}
	\vspace{-0.1cm}
\end{figure}

In the BF$A$'s RDF, the four base ISAs each became an RSE. 
The RDF used a feature called {\tt isDistinctBaseISA} to note
the four base ISAs. The information was represented implicitly,
resulting in use of different RSEs and a more compact RDF. 

For the second paragraph shown in Figure~\ref{figApp4RSE-CO},
the main subject was the term {\tt IALIGN}. The BF$\phi$'s RDF
captured this main subject correctly. 
However, the second sentence in the paragraph mentioned that
``compressed ISA extension relaxes IALIGN to 16 bits''. 
This fact was not captured in the BF$\phi$'s RDF.
Instead, it used a more generic relation {\tt hasISAExtension}
to relate {\tt IALIGN} to {\tt compressedISAExtension}. 

The BF$A$'s RDF captured the second fact by creating the
subject entity {\tt compressedISAExtension} and using 
the relation {\tt relaxes} to relate the entity to
{\tt IALIGN}. The {\tt IALIGN} was still kept as a subject
entity. However, because it appeared in the object field
of a triple, {\tt IALIGN} was no longer counted as
a root subject entity. 
It was also interesting to notice that in the BF$A$'s RDF
the second Fact explicitly enumerated a few example values that
the {\tt IALIGN} cannot take on. There was nothing incorrect by
adding this extra information to the RDF. 

Figures~\ref{figAppP63a} 
and Figures~\ref{figAppP63b} then show the comparison for the
third paragraph in Figure~\ref{figApp4RSE-CO}. 
The main subject of this paragraph was about ``invisible trap''. 
The paragraph provided three example cases where this type of trap
can take places: missing instructions, non-resident page faults,
and device interrupts. Comparing BF$\phi$'s RDF to
BF$A$'s RDF, we observed that they both included the entities 
to represent the three examples. However, the entities in the BF$\phi$'s RDF
attached prefixes ``emulating'' and ``handling'' to the entities
while the BF$A$'s RDF just used the three example names as given. 
The RSE of the BF$\phi$'s RDF was {\tt Trap} (we might see it as an instance)
which was declared as an {\tt InvisibleTrap}. In contrast, the three example cases were
all treated as RSEs in the BF$A$'s RDF, which were all declared
as a subclass of {\tt InvisibleTrap}. 
Then, the {\tt InvisibleTrap} was declared as a {\tt Trap} 
(we might see it as a concept) which
is more intuitive. Overall, we observed that the BF$A$'s RDF
was easier to interpret than the BF$\phi$'s RDF. 

With the BFs provided, the BF$A$'s RDF further connected
{\tt Trap} to {\tt EEI} and provided a Fact on the 
{\tt ExecutionEnvrionment} that was connected to 
{\tt EEI} and {\tt hart}. These connections were useful when
we merged this paragraph-specific KG with the KGs 
from other paragraphs that provided more detail on 
the EEI and hart.  

The fourth example is shown in Figures~\ref{figAppP81}. 
The BF$\phi$'s RDF used {\tt RV32E} as the RSE while
the BF$A$'s RDF used {\tt Chapter4} as the RSE. 
Both RDFs contained the same information from the paragraph.
The BF$A$'s RDF had extra information on RV32I due to the
BFs provided. 

\subsection{Two Additional Interesting Cases}
\label{App.06}

In this section, we discuss two more cases in detail to
highlight the effect from BF$\phi$ to BF$A$. These two cases
were among the more challenging cases we had encountered. 
The BF$\phi$'s RDFs did not pass the entailment check
and we had to carefully select the BFs to obtain 
a satisfactory RDF with BF$A$.

\begin{figure}[h]
	\centering
	\includesvg[inkscapelatex=false, width=3in]{Figures/figAppP79.1a.svg}
	\vspace{-0.3cm}
	\caption{\small The paragraph for the first example}
	\label{figAppP79.1a}
	\vspace{-0.3cm}
\end{figure}

\begin{figure}[H]
	\centering
	\includesvg[inkscapelatex=false, width=3in]{Figures/figAppP79.1b1.svg}
	\vspace{-0.3cm}
	\caption{\small The main body of the RDF created by
	BF$\phi$ for the paragraph in Figure~\ref{figAppP79.1a}}
	\label{figAppP79.1b1}
	\vspace{-0.1cm}
\end{figure}

\begin{figure}[h]
	\centering
	\includesvg[inkscapelatex=false, width=3in]{Figures/figAppP79.1b2.svg}
	\vspace{-0.3cm}
	\caption{\small The remaining RDF continuing from 
		Figure~\ref{figAppP79.1b1}}
	\label{figAppP79.1b2}
	\vspace{-0.1cm}
\end{figure}

The paragraph for the first example is shown in Figure~\ref{figAppP79.1a}. 
This paragraph was challenging because it involved several specific
concepts. Take the first phrase ``avoid intermediate instruction
sizes'' as an example. This phrase can include the concept
``instruction sizes'', the concept ``intermediate instruction 
sizes'' and the concept ``avoid intermediate instruction sizes''. 
Choosing which concept to start can affect the remaining KGC process. 

The RDF from the BF$\phi$ is shown in Figure~\ref{figAppP79.1b1}
continuing in Figure~\ref{figAppP79.1b2}. As seen, this RDF defined
various verbs (``adopted'', ``avoid'', ``helps'', and so on) as
rdf property and used them as relations to connect the various entities. 
These property definitions all failed the entailment check. 
The RDF even created an entity called {\tt wanted} to capture the
fact that ``intermediate instruction sizes'' were wanted to be avoided
(by the RISC-V ISA developer).

\begin{figure}[h]
	\centering
	\includesvg[inkscapelatex=false, width=3in]{Figures/figAppP79.1c1.svg}
	\vspace{-0.1cm}
	\caption{\small The main body of the RDF created by
		BF$A$ for the paragraph in Figure~\ref{figAppP79.1a}}
	\label{figAppP79.1c1}
	\vspace{-0.1cm}
\end{figure}

\begin{figure}[h]
	\centering
	\includesvg[inkscapelatex=false, width=3in]{Figures/figAppP79.1c2.svg}
	\vspace{-0.1cm}
	\caption{\small The remaining RDF continuing from 
		Figure~\ref{figAppP79.1c1}}
	\label{figAppP79.1c2}
	\vspace{-0.1cm}
\end{figure}

The BFs provided to BF$A$ included three facts specifying that
``avoid intermediate instruction size'', ``base hardware implementation'',
and ``larger number of integer registers'' all should be treated
as a single concept. The result RDF is shown in Figure~\ref{figAppP79.1c1}
(continuing in Figure~\ref{figAppP79.1c2}) where the three were created
as subject entities, in addition to the 
subject entity {\tt 32\_bit\_instruction\_size}
whose equivalent form also appeared in the BF$\phi$ RDF before
(Figure~\ref{figAppP79.1b1}). 

In the BF$A$'s RDF, instead of treating verbs as rdf property
like that in the the BF$\phi$ RDF (Figure~\ref{figAppP79.1b2}), three
verbs ({\tt supports}, {\tt helps\_performance},
and {\tt simplifies}) were defined with ``rdfs1:label''.  
Then, they were used to connect the four subject entities,
which more accurately reflected the information conveyed by the paragraph. 

\subsubsection{The Second Example}
\label{App.06.1}

Figure~\ref{figAppP20.3a} shows the paragraph of the second
example. The BF$\phi$'s RDF is shown in Figure~\ref{figAppP20.3b1}
continuing in Figure~\ref{figAppP20.3b2}.

\begin{figure}[h]
	\centering
	\includesvg[inkscapelatex=false, width=3in]{Figures/figAppP20.3a.svg}
	\vspace{-0.3cm}
	\caption{\small The paragraph for the second example}
	\label{figAppP20.3a}
	\vspace{-0.1cm}
\end{figure}

Similar to the previous example, this paragraph contained
phrases where multiple choices of entities might be extracted. 
For example, the phrase ``increasing address space size'' might
be represented by the phrase itself, ``address space'',
Or ``address space size''.  As seen in Figure~\ref{figAppP20.3b1},
BF$\phi$ chose the long names
{\tt increasing\_address\_space\_size} and
{\tt supporting\_running\_existing\_binaries},
and treated them as rdf property (Figure~\ref{figAppP20.3b2}). 

The RDF in Figure~\ref{figAppP20.3b1} contained two Facts, one
for mips and the other for sparc. The two essentially
Were represented in the same way. It was interesting to see that
each Fact contained duplicated comments and labels, one with
rdf namespace and the other with rdfs1 namespace. 
The texts with the comments and labels were additional
information provided by GPT3.5 itself. Again, this shows that
GPT3.5 has the ability to provide its own background facts. 

\begin{figure}[h]
	\centering
	\includesvg[inkscapelatex=false, width=3in]{Figures/figAppP20.3b1.svg}
	\vspace{-0.3cm}
	\caption{\small The main body of the RDF created by
		BF$\phi$ for the paragraph in Figure~\ref{figAppP20.3a}}
	\label{figAppP20.3b1}
	\vspace{0.3cm}
\end{figure}

\begin{figure}[h]
	\centering
	\includesvg[inkscapelatex=false, width=3in]{Figures/figAppP20.3b2.svg}
	\vspace{-0.3cm}
	\caption{\small The remaining RDF continuing from 
		Figure~\ref{figAppP20.3b1}}
	\label{figAppP20.3b2}
	\vspace{-0.1cm}
\end{figure}

The three property-based Facts in Figure~\ref{figAppP20.3b2} did not
pass the entailment check. While BF$\phi$'s RDF treated mips and
sparc as two main subjects, it did not capture what
``strict superset policy'' meant other than it is a property used
to relate the ISA to  
{\tt supporting\_running\_existing\_binaries}. 

After analyzing the paragraph more carefully, we found that the
main topic was about the ``strict superset policy''. MIPS and
SPARC were just examples that adopt this policy. The BF$\phi$
did not capture this point. 
Consequently, in BF$A$ we added a background fact to state that
"strict superset policy" refered to a choice of ISA design.
In addition, we also added a background fact about the
``address space''. 
The resulting RDF is shown in Figure~\ref{figAppP20.3c}. 
This RDF is more compact than the BF$\phi$'s RDF shown before. 

\begin{figure}[h]
	\centering
	\includesvg[inkscapelatex=false, width=2.5in]{Figures/figAppP20.3c.svg}
	\vspace{-0.1cm}
	\caption{\small The BF$A$'s RDF for the paragraph in Figure~\ref{figAppP20.3a}}
	\label{figAppP20.3c}
	\vspace{-0.1cm}
\end{figure}

It was interesting to notice that instead of treating
"strict superset policy" as an entity, BF$A$ used the entity
{\tt superset\_policy}. It was declared as a type of
{\tt DesignPolicy} based on the background fact we provided. 
Instead of using the longer entity 
{\tt increasing\_address\_space\_size} as before,
it now uses a simpler (and more generic) entity
 {\tt address\_space} as provided in our BFs. 
Note that ``address space'' was a concept frequently mentioned
in other parts of the Spec. Hence, using this entity can help
make connections to the KGs of other paragraphs later. 

The two main facts for mips and sparc essentially said that
the ISA had a design policy called ``superset policy'' that
involved using 64-bit address space to support 32-bit
binaries. This representation reflected the information
conveyed by the paragraph better than the BF$\phi$'s RDF
shown before.  

The two examples discussed in this section can be seen as
two representative cases that are opposite to each other. 
The first example shows that 
BF$A$ moves from using shorter names as that in BF$\phi$
to using longer names, resulting in more complex RDF and 
uncovering more relations (i.e. more coverage). 
In the second example,
BF$A$ moves from
using longer names to shorter names, resulting in more
compact RDF by focusing on the right topic
and the use of more generic entities (i.e.
less coverage but more accurate). 

In general, the changes from BF$\phi$ to BF$A$ can be diverse.
These changes not only depend on how the BFs influence the
GPT3.5's KGC behavior, but also depend on the paragraph being
processed.

\begin{figure*}[htb]
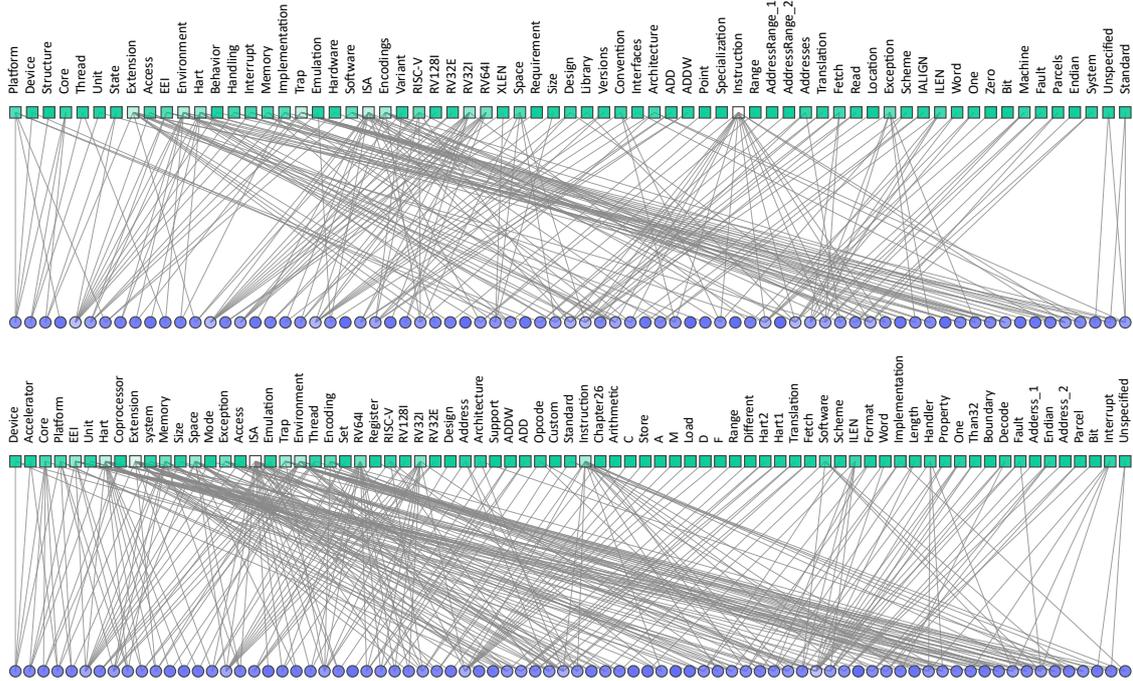

	\centering
	\includesvg[inkscapelatex=false, width=6.5in]{Figures/figApp01.svg}
	\includesvg[inkscapelatex=false, width=6.5in]{Figures/figApp02.svg}
	\vspace{-0.5cm}
	\caption{\small \textbf{Top} (BF$\phi$): 
		67 subject concepts shared by at least the RDFs of two paragraphs with 
		the total number of 320 edges shown in the bipartite graph ; Note that this result includes
		all RDFs even though they fail the entailment check; \textbf{Bottom} (BF$A$): 75 concepts shared by at least two paragraph and the total number of edges increases to 454; In the bipartite graphs, the bottom dots each represents a paragraph from chapter 1 and the upper squares each represents a subject concept. More transparent the color indicates more edges are connected between the concepts and paragraphs.}
		\label{figApp01} 
		\vspace{-0.2cm}
\end{figure*}

\subsection{Cross-Paragraph Connections}
\label{App.07}

A total of 433 and 452 subject entities were extracted by 
BF$\phi$ for chapters 1 and 2, respectively (disregarding the entailment
check results). In contrast, a total of 597 and 577 subject entities were extracted by 
BF$A$ for chapters 1 and 2, respectively. 

To show how many paragraphs can be connected through the
subject entities, we first used a simple method to group  
subject entities into high-level subject concepts. 
If subject entities shared the same suffix word, we grouped them together.
For example, ``CSRInstruction" and ``StoreInstruction" were grouped with
the high-level concept ``Instruction". 
Suffix can be easily split from the entity phrase since RDF
already formated the phrases into Camel or Snake case. 
We further used a stemming tool \cite{nltk-stemmer} to remove morphological 
affixes from the suffix words so that words with the same stem would be 
grouped together. For example, 
``encodings" and ``encodes" belonged to the same group. 

For chapter 1, we collected a total of 168 and 178 subject concepts 
from BF$\phi$ and BF$A$, respectively. 
These subject concepts provided in total 
421 connections for BF$\phi$ and 557 connections for BF$A$, respectively,
to all the paragraphs.   


Figure \ref{figApp01} shows two bipartite graphs between 
subject concepts and paragraphs, one for BF$\phi$ and
the other for BF$A$. An edge means the subject concept
appears in the paragraph. Only those subject concepts that connect
at least two paragraphs are shown in the graphs. 
For BF$\phi$, the graph has 67 subject concepts with 320 edges.
For BF$A$, the graph has 75 subject concepts with 454 edges. 
In other words, BF$A$'s RDFs have more cross-paragraph connections
through the subjects concepts. 

\begin{figure*}[htb]
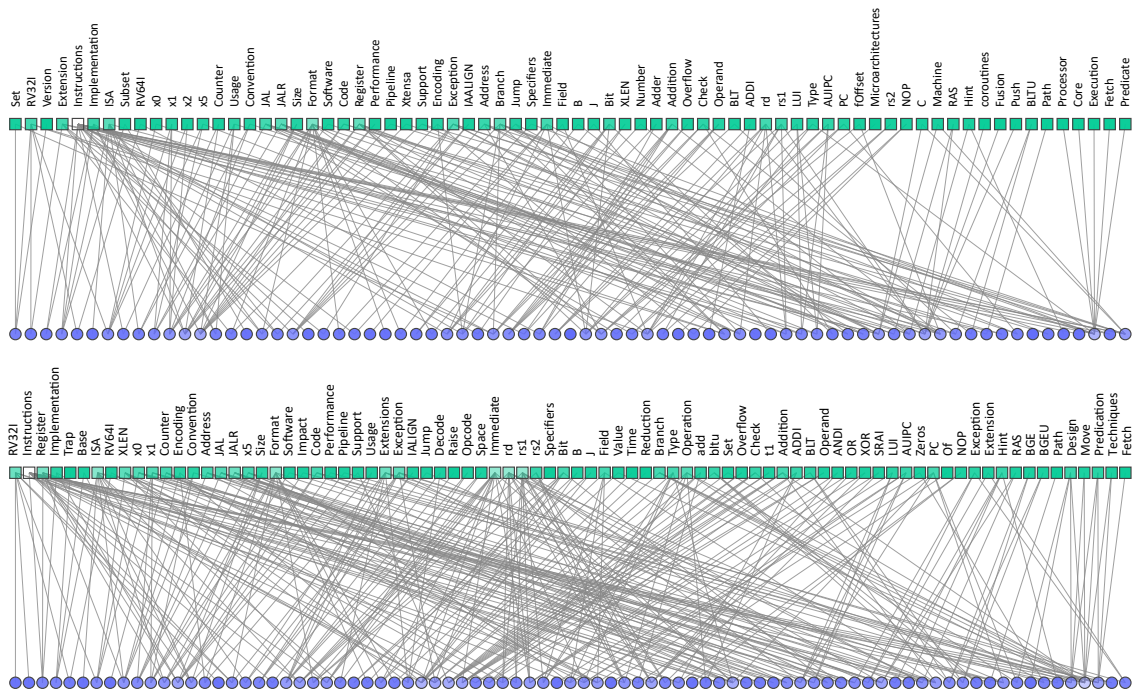

	\centering
	\includesvg[inkscapelatex=false, width=6.5in]{Figures/figApp03.svg}
	\includesvg[inkscapelatex=false, width=6.5in]{Figures/figApp04.svg}
	\vspace{-0.5cm}
	\caption{\small \textbf{Top} (BF$\phi$): 72 subject concepts shared by at least the RDFs of two paragraphs with 
		the total number of 307 edges shown in the bipartite graph; Note that this result includes
		all RDFs even though they fail the entailment check; \textbf{Bottom} (BF$A$): 82 concepts shared by at least two paragraph and the total number of edges increases to 388; In the bipartite graphs, the bottom dots each represents a paragraph from chapter 2 and the upper squares each represents a subject concept.}
	\label{figApp03} 
	\vspace{-0.2cm}
\end{figure*}

For chapter 2, we collected a total of 211 and 191 subject concepts 
from BF$\phi$ and BF$A$, respectively. 
These subject concepts provided in total 
446 connections for BF$\phi$ and 497 connections for BF$A$, respectively,
to all the paragraphs.
Figure \ref{figApp03} shows similar bipartite graphs. 
For BF$\phi$, the graph has 72 subject concepts with 307 edges.
For BF$A$, the graph has 82 subject concepts with 388 edges. 
Again, BF$A$'s RDFs have more connections. 

Table \ref{tableApp01} shows the top five subject concepts 
that have the highest
connectivity (the highest number of occurrences) 
in Figure \ref{figApp01} (top). 
The subject entities grouped
under the same concept are also listed. 
Table \ref{tableApp02} shows the concepts and the corresponding entities
for Figure \ref{figApp01} (bottom).
Table \ref{tableApp03} shows the concepts and the corresponding entities
for Figure \ref{figApp03} (top).
Table \ref{tableApp04} shows the concepts and the corresponding entities
for Figure \ref{figApp03} (bottom).

\begin{table*}[htb]
	\centering
	\begin{tabular}{c|l}
		\textbf{Subject Concept} & \textbf{BF$\phi$ Subject Entities}    \\ \hline
		Instruction (23)  &  New\_Instructions, cacheControlInstruction, \\
		& fenceInstruction, StoreInstruction, IllegalInstructions\\
		& specificInstructions, Variable\_Length\_Instructions, \\
		&  32\_bit\_instruction, Optional\_Longer\_Instructions,  \\
		&  LittleEndianInstruction, ErrorfulInstruction, \\
		& unprivilegedInstructions, Instruction, MachineInstruction,\\
		& optional\_compressed\_instruction, LoadInstruction \\ \hline
		Trap (19) & Opcode\_Traps, Trap, InvisibleTrap, RequestedTrap, \\
		& FatalTrap, ContainedTrap \\ \hline
		ISA (19) & RISC-V\_ISA, Base\_Integer\_ISA, Base\_ISA, singleISA, ISA, \\
		& whyNotSingleISA, Unprivileged\_ISA \\ \hline 
		Extension (16) & Additional\_Instruction\_Set\_Extension, \\
		& Specialized\_Instruction\_Set\_Extension, \\
		& Optional\_Extension, nonConformingExtension, \\
		& InstructionSetExtensions, Subsequent\_Extensions, \\
		& GC\_Extensions, Standard\_Compressed\_ISA\_Extension, \\
		& StandardExtensions, StandardInstructionSetExtension, \\
		& compressed\_instruction\_set\_extensions, \\
		& OtherExtensions, nonStandardExtension, Extensions\\ \hline
		Execution Environment (15) & Software\_Execution\_Environment, \\
		& Hardware\_Execution\_Environment, \\
		& RISC-V\_Execution\_Environment, \\
		& Hardware\_and\_Software\_Execution\_Environment, \\
		& BareMetalEnvironment, Execution\_Environment\\
		& OperatingSystemEnvironment \\ \hline

	\end{tabular}
	\caption{\small BF$\phi$ Entities from Chapter 1 with top occurrences. 
		The first column shows the domain suffix term (subject concept) that is shared by the entities.   
		The number in parenthesis indicates the number of paragraphs where the term occurs.}
	\label{tableApp01}
\end{table*}

\begin{table*}[htb]
	\centering
	\begin{tabular}{c|l}
		\textbf{Subject Concept} & \textbf{BF$\phi$ Subject Entities}    \\ \hline
		Extension (38) & Instruction\_Set\_Extension, Extension, nonStandardExtension,\\
		& vendorSpecificNonStandardExtension, compressed\_extension, \\
		& standard\_GC\_extensions, Compressed\_ISA\_Extension, \\
		& customExtension, StandardCompressedInstructionExtension,\\ 
		& StandardLargerThan32BitInstructionSetExtension, \\
		& NonConformingExtension, StandardFloatingPointExtension, \\
		& openToExtensions, ISA\_Extension, \\
		& OtherExtensions, has\_extension \\ \hline
		ISA (33) & ISA, RISC-V\_ISA, has\_base\_ISA, Base\_Integer\_ISA, unifiedISA, \\ 
		& not\_treating\_design\_as\_single\_ISA, Unprivileged\_ISA, \\
		& standard\_IMAFD\_ISA, singleISA, notSingleISA \\ \hline
		Trap (33) & Trap, differences\_in\_addressing\_and\_illegal\_instruction\_traps, \\
		& illegal\_instruction\_traps, NoTrap, FatalTrap, ContainedTrap, \\
		& RequestedTrap, DefinedSolelyToCauseRequestedTraps,  \\
		& InvisibleTrap, OpcodeTrap \\ \hline
		Instrucion (31) & RISCVBaseInstructions, MultiplicationInstruction, \\
		& DoublePrecisionInstruction, ControlFlowInstruction, \\
		& IntegerComputationalInstruction, SinglePrecisionInstruction, \\
		& DivisionInstruction, AtomicInstruction, CompressedInstruction, \\
		& LoadInstruction, Instruction, MachineInstruction\\
		& Compressed\_16\_Bit\_Instruction, 32\_Bit\_Instruction, \\
		& illegal\_30\_bit\_instructions, illegalInstruction, 16BitInstruction, \\
		& AllZeroBitsInstruction, Variable-Length\_Instruction, \\
		& AllOnesInstruction, KnownErrorfulInstructions, StoreInstruction, \\
		& MissingInstruction, UnprivilegedInstructions, CommonInstruction\\ \hline
		Execution Environment (15) & HardwareAndSoftwareExecutionEnvironment, \\
		& SoftwareExecutionEnvironment, \\
		& SupervisorLevelExecutionEnvironment, \\
		& UserLevelExecutionEnvironment, OuterExecutionEnvironment, \\
		& BareMetalEnvironment, ExecutionEnvironment \\ \hline 
		Hart (15) & Hart, HostHart, GuestHart, EachHart, AllHarts \\ \hline 
	\end{tabular}
	\caption{\small BF$A$ Entities from Chapter 1 with top occurrences. }
	\label{tableApp02}
\end{table*}

\begin{table*}[htb]
	\centering
	\begin{tabular}{c|l}
		\textbf{Subject Concept} & \textbf{BF$\phi$ Subject Entities}    \\ \hline
		Instruction (23) & CSRInstructions, systemInstructions, CurrentInstruction, 16BitInstructions, \\
		& Jump\_Instruction, Branch\_Instruction, Conditional\_Branch\_Instruction, \\
		& Instruction, LoadUpperImmediateInstruction, RegularInstruction, \\
		& hasInstruction, IntegerComputationalInstructions, \\
		& SingleAdditionalBranchInstruction, AdditionalInstructions, \\
		& ControlTransferInstructions, unconditionalJumpInstructions, \\
		& ConditionalBranchInstruction, 16BitAlignedInstructions, \\
		& AddressOfBranchInstruction, PredicatedInstructions \\ \hline
		Format (19) & InstructionFormat, 3AddressFormat, 2AddressFormat, S\_Format, R\_Format, \\
		& U\_Format, I\_Format, B\_InstructionFormat, J\_InstructionFormat, B\_format,\\
		& Base\_Instruction\_Formats, usesFormat, JTypeFormat, BTypeInstructionFormat \\ \hline
		Register (12) & numberOfRegisters, 32IntegerRegisters, IntegerRegisters, hasDestinationRegister, \\
		& placesResultInRegister, AlternateLinkRegister, RegularReturnAddressRegister,\\
		& GeneralPurposeRegister, linkRegister, TwoRegisters \\ \hline
		ISA (9) & baseIntegerISA, integerISA, completeISA, RV32I\_ISA, BaseISA, ISA \\ \hline 
		Extension (9) & OtherISAExtensions, InstructionSetExtension, NonConformingExtension, \\
		& CompressedInstructionSetExtension, AExtension, Extension \\ \hline 
	\end{tabular}
	\caption{\small BF$\phi$ Entities from Chapter 2 with top occurrences.}
	\label{tableApp03}
\end{table*}

\begin{table*}[htb]
	\centering
	\begin{tabular}{c|l}
		\textbf{Subject Concept} & BF$A$ \textbf{Subject Entities}    \\ \hline
		Instruction (23) & CSRInstructions, ConditionalBranchInstruction, BranchInstruction, \\
		& reservedInstruction, regular\_instruction, load\_upper\_immediate\_instruction, \\
		& IntegerComputationalInstruction, JumpInstruction, AdditionalInstructions, \\
		& LoadStoreInstructions, ControlTransferInstructions, \\
		& unconditionalJumpInstructions, PredicatedInstructions, \\ \hline
		Immediate (17) & 5-bit\_Immediate, Immediate, J\_immediate, J\_shifted\_immediate, \\ 
		& 20\_bit\_immediate, U\_immediate, U\_shifted\_immediate \\ \hline 
		ISA (16) & SubsetOfBaseIntegerISA, ISA, BaseIntegerISA, RISC-V\_ISA, \\
		& Unprivileged\_State\_for\_Base\_Integer\_ISA, Complete\_Set\_of\_Base\_Integer\_ISA, \\ 
		& integer\_ISA, baseISA \\ \hline 
		Register (15) & CSRRegisters, hasNoStackPointerOrSubroutineReturnAddressLinkRegister, \\
		& usesRegisterX5AsAlternateLinkRegister, 16\_registers, number\_of\_registers, \\
		& larger\_number\_of\_integer\_registers, integer\_register, \\
		& frequently\_accessed\_registers, returnAddressRegister \\ \hline 
		Format (12) & InstructionFormat, 2-address\_format, OptionalCompressed16BiatInstructionFormat, \\ 
		& CoreInstructionFormat, B\_format, S\_format, U\_format, J\_format, \\
		& BaseInstructionFormats, BTypeInstructionFormat, alternateLinkRegister, \\
		& ReturnAddressRegister \\ \hline 
	\end{tabular}
	\caption{\small BF$A$ Entities from Chapter 2 with top occurrences.}
	\label{tableApp04}
\end{table*}

\subsection{Background Facts}
\label{App.08}

There were 204 background facts accumulated in our experiment.
They are shown in Figures~\ref{figApp05}-\ref{figApp07} at the end. 
Note that some of them 
were collected after their first appearance (recall that we processed
paragraphs one by one in the order as they appeared in the Spec). 
For example, for some
instruction names their BFs were not required for processing the 
paragraphs that defined them. The BFs were collected after the
these paragraphs were processed and were used to process later
paragraphs referencing the instruction names. 

\begin{figure*}[htb]
	\centering
	\includesvg[inkscapelatex=false, width=6in]{Figures/figApp05.svg}
	\vspace{-0.2cm}
	\caption{\small Background facts}
	\label{figApp05} 
	\vspace{-0.2cm}
\end{figure*}

\begin{figure*}[htb]
	\centering
	\includesvg[inkscapelatex=false, width=6in]{Figures/figApp06.svg}
	\vspace{-0.2cm}
	\caption{\small Background facts Cont'd}
	\label{figApp06} 
	\vspace{-0.2cm}
\end{figure*}

\begin{figure*}[htb]
	\centering
	\includesvg[inkscapelatex=false, width=6in]{Figures/figApp07.svg}
	\vspace{-0.2cm}
	\caption{\small Background facts Cont'd}
	\label{figApp07} 
	\vspace{-0.2cm}
\end{figure*}

\end{document}